\documentclass[10pt,twocolumn,letterpaper]{article}
\usepackage[pagenumbers]{cvpr}

\usepackage{booktabs}
\usepackage{makecell}
\usepackage{pifont} 
\usepackage{xcolor} 
\usepackage{colortbl}  %
\usepackage{multirow}
\usepackage{multicol}

\definecolor{cvprblue}{rgb}{0.21,0.49,0.74}
\usepackage[pagebackref,breaklinks,colorlinks,allcolors=cvprblue, urlcolor=magenta]{hyperref}

\title{\parbox{0.05\textwidth}{\includegraphics[width=\linewidth]{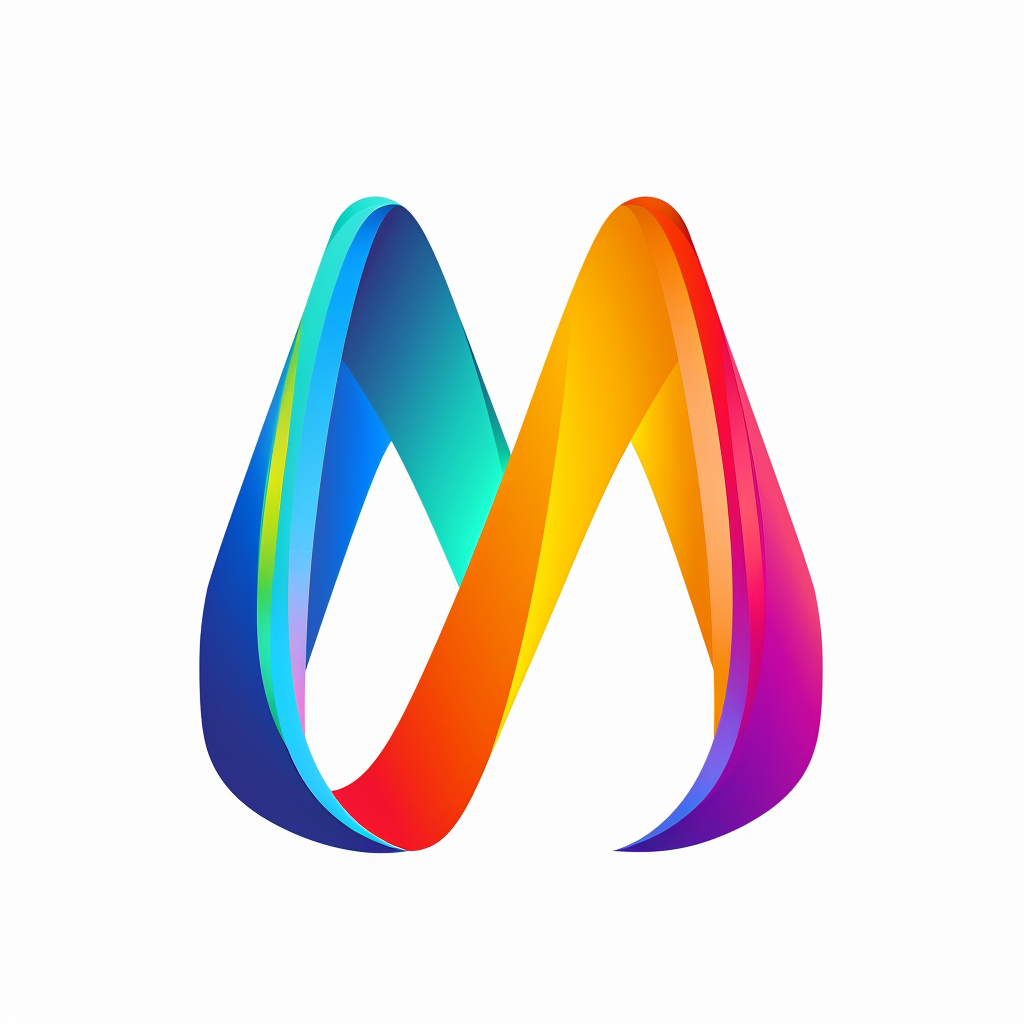}} MotionShop: Zero-Shot Motion Transfer in Video Diffusion Models with Mixture of Score Guidance}

\author{Hidir Yesiltepe \qquad Tuna Han Salih Meral \qquad Connor Dunlop \qquad Pinar Yanardag \\
Virginia Tech \\
\tt \small \{hidir, tmeral, cdunlop, pinary\}@vt.edu \\
\small \url{motionshop-diffusion.github.io}
}

\begin{document}

\twocolumn[{
\maketitle
\begin{center}
    \captionsetup{type=figure}
    \vspace{-1em}
\newcommand{\imwidth}{1\textwidth}

\begin{tabular}{@{}c@{}}

\parbox{\imwidth}{\includegraphics[width=\imwidth]{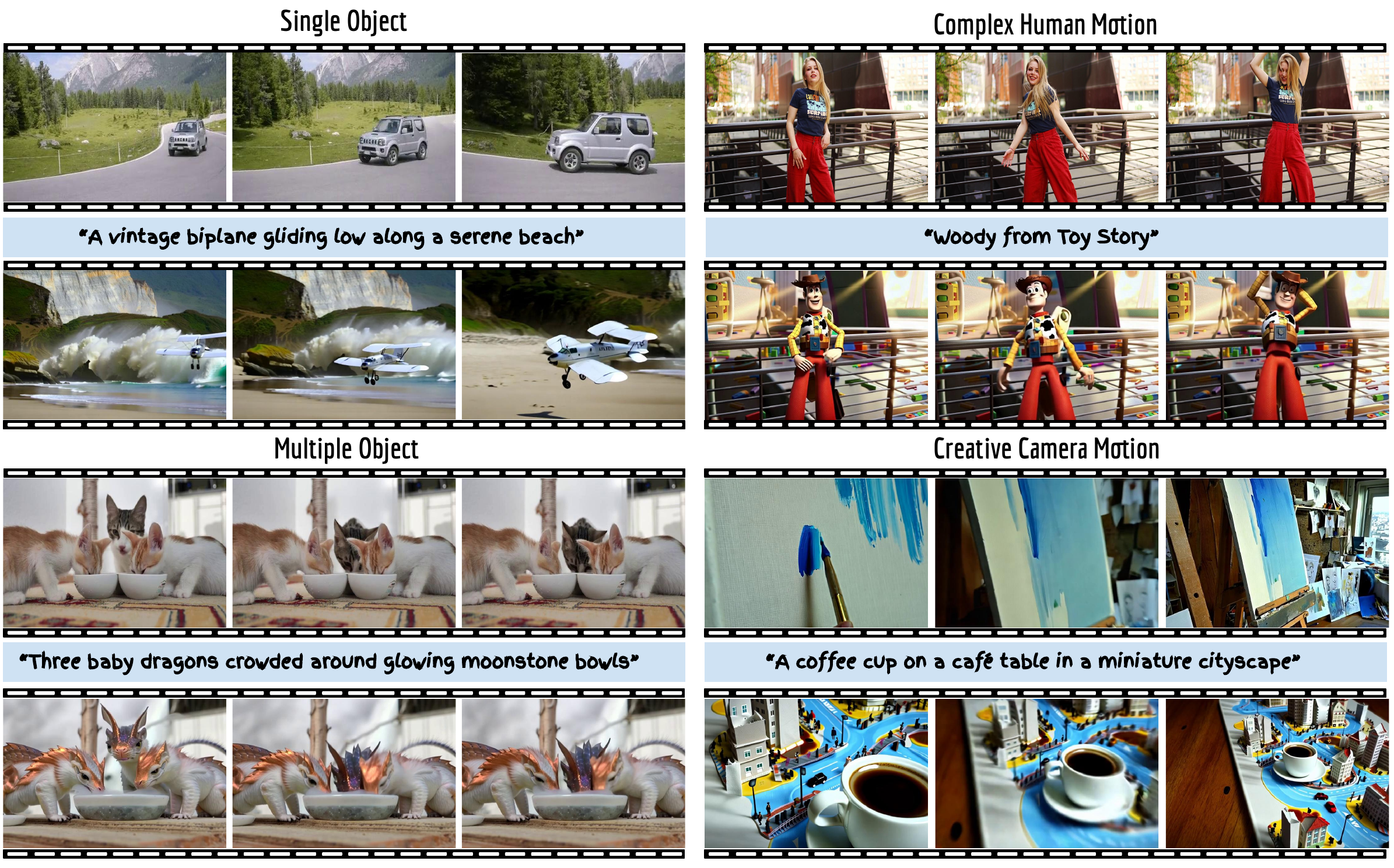}}
\\

\vspace{1em}
\end{tabular}
    \vspace{-2.5em}
    \captionof{figure}{Mixture of Score Guidance (MSG), a novel approach for zero-shot motion transfer in diffusion models, enables high-fidelity motion synthesis across diverse scenarios. MSG successfully handles various motion patterns including complex object movements and camera trajectories. Full video results are available in the supplementary material.} 
    \label{fig:teaser}
\end{center}
}]

\maketitle
\begin{abstract} 
In this work, we propose the first motion transfer approach in diffusion transformer through Mixture of Score Guidance (MSG), a theoretically-grounded framework for motion transfer in diffusion models. Our key theoretical contribution lies in reformulating conditional score to decompose motion score and content score in diffusion models. By formulating motion transfer as a mixture of potential energies, MSG naturally preserves scene composition and enables creative scene transformations while maintaining the integrity of transferred motion patterns. This novel sampling operates directly on pre-trained video diffusion models without additional training or fine-tuning. Through extensive experiments, MSG demonstrates successful handling of diverse scenarios including single object, multiple objects, and cross-object motion transfer as well as complex camera motion transfer. Additionally, we introduce MotionBench, the first motion transfer dataset consisting of 200 source videos and 1000 transferred motions, covering single/multi-object transfers, and complex camera motions.
\end{abstract}
    
\section{Introduction}
\label{sec:intro}

Diffusion-based video generation models have gained substantial attention for their ability to produce high-quality, diverse video content. These models, driven by advances in text-to-video generation, open new possibilities for automated and creative video synthesis \cite{blattmann2023stable, hong2022cogvideo, guo2023animatediff, make-a-video, wang2023modelscope, yu2024efficient, yang2024cogvideox}. Motion transfer in generative models \cite{jeong2024vmc, chen2023motion, zhang2023motioncrafter, zhao2025motiondirector, wang2024motion, yatim2024space}, has become a significant research area, focusing on transferring the motion from one video to another, often guided by text prompts.  Consider the complex transformation depicted in Fig. \ref{fig:teaser}, where a ground vehicle's trajectory is reimagined as the flight path of an aircraft. Such motion transfer  involves more than merely replacing the car with a plane. For instance, translating the movement of a car into a plane gliding over a beach, as described by the text prompt (see Fig. \ref{fig:teaser}) requires a significant adjustment in environmental context. This includes transforming how the car interacts with the road to how an aircraft engages with the sky.  This level of control is particularly important as it enables users to create videos with motions that are   challenging to describe through text prompts alone, such as complex camera motions (see Fig. \ref{fig:teaser}). 

Recent video generation and editing methods have focused on disentangling motion and appearance characteristics. Various approaches have emerged: MotionDirector~\cite{zhao2025motiondirector} uses an appearance-debiased temporal loss with dual-path LoRA architecture, while DreamVideo~\cite{wei2024dreamvideo}, Customize-A-Video~\cite{ren2024customize}, and MotionCrafter~\cite{zhang2023motioncrafter} employ dedicated processing branches. VMC~\cite{jeong2024vmc} combines fine-tuning and inversion techniques targeting temporal layers, and DMT~\cite{yatim2024space} leverages space-time feature loss using DDIM inversion and UNet activations. Motion Inversion ~\cite{wang2024motion} uses motion embeddings trained from reference videos for temporal dynamics control. Despite these advancements,  motion control in video generation remains challenging because of the complex interplay between spatial and temporal dimensions in video. Controlling motion is essential for applications in entertainment, advertising, and virtual reality, where specific and consistent movements are crucial to communicate a narrative or aesthetic vision.
\begin{figure}[t!]
    \centering
    \begin{tabular}{c}
        \hspace{-0.3cm}
        \includegraphics[width=\linewidth]{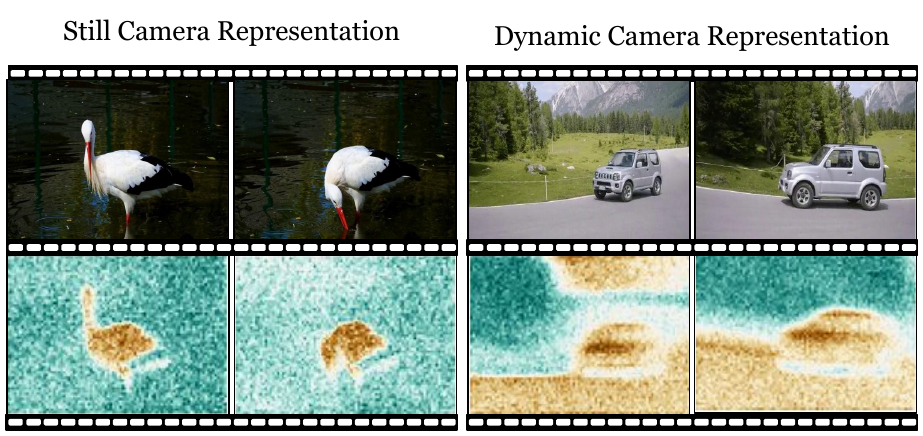} \\
    \end{tabular}
    \caption{\textbf{Our intuition.} Visualization of motion characteristics $\mathcal{M}(z)$ extracted from early-timestep conditional scores. \textbf{(Left)} Multiple object motion representation showing the simultaneous movement of two objects. \textbf{(Right)} Combined object and camera motion representation demonstrating how our method captures both local object motion and global camera movement patterns. The visualizations are obtained from the conditional score maps $\nabla_{z_t} \log p_t(z|y)$ at early timesteps $t \ll T$.}
    \label{fig:motion_representation}
\end{figure}

However, while these methods are effective in straightforward motion transfer tasks involving single objects without significant background or object transformations, they struggle with more challenging motion transfer tasks. They often fail to adequately transform the scene, merely replacing one object with another without aligning the scenery  with the changes specified in the text prompt. Other methods may dramatically alter the scene without preserving the original motion.On the other hand, video editing methods  \cite{kara2024rave, geyer2023tokenflow, ceylan2023pix2video, qi2023fatezero} utilize structural similarities between source and target videos.  However, their performance is limited when it comes to multi-object motion transfer or managing complex camera movements. Additionally, these methods struggle with significant shape transformations, such as turning a car into a flying plane (see Fig. \ref{fig:teaser}). These limitations highlight the need for more advanced motion transfer techniques that can handle a wider range of transformations and motions without being constrained by the physical similarities between the source and target videos. Such capabilities would significantly enhance the flexibility and applicability of generative models in video editing and animation, opening up new possibilities for creative and practical applications.

In this paper, we present Mixture of Score Guidance (MSG), a novel approach for motion transfer in diffusion-based generative video models. Our method builds upon a novel conditional score reformulation, where we formulate motion transfer as a mixture of potential energies in the score space of diffusion models. By leveraging the observation that reformulated conditional scores encode rich motion information in early diffusion timesteps, MSG successfully isolates and transfers motion patterns. We establish the mathematical connection between score mixing and Langevin dynamics, providing theoretical perspectives for stable motion transfer. Through extensive experimentation, we demonstrate that MSG enables high-fidelity motion transfer across diverse scenarios without requiring model fine-tuning or additional training data. Our work extends beyond single-motion cases to handle multiple motion sources, and complex camera motions offering a unified approach to video motion transfer. Our contributions are as follows:

\begin{itemize}
    \item We introduce Mixture of Score Guidance (MSG), a theoretically grounded framework for motion transfer that formulates the problem through the lens of statistical mechanics. Our method operates directly in score space without requiring additional training or fine-tuning.
    
    \item We demonstrate the relationship between conditional scores and motion information, showing that score mixing in early diffusion steps provides an effective approach to motion transfer.
    
    \item We show that MSG's theoretical foundations naturally extend to complex scenarios including multi-motion synthesis and complex camera motion transfer.
    
    \item We introduce MotionBench, a comprehensive motion transfer benchmark comprising 200 diverse source videos and 1000 transferred sequences. This benchmark spans single/multi-object transfers,and  camera motion variations enabling systematic evaluation of motion transfer methods across a broad range of scenarios.

\end{itemize}

\section{Related Work}
\label{sec:related}

\subsection{Text-to-Video Generation}
Transformer architectures have emerged as a powerful foundation for video generation tasks. Early research scaling transformers for T2V applications, including Sora \cite{openai2024sora}, CogVideo \cite{hong2022cogvideo}, CogVideoX \cite{yang2024cogvideox} and LATTE \cite{ma2024latte}, established the viability of this approach. The introduction of Diffusion Transformers \cite{peebles2023scalable} further cemented transformers as core components in video diffusion models. Several works have introduced specialized conditioning inputs: ControlVideo \cite{zhang2023controlvideo} leverages depth maps, DragNUWA \cite{yin2023dragnuwa} employs motion trajectories, while VideoDirectorGPT \cite{lin2023videodirectorgpt} and related approaches \cite{lian2023llm, chen2023motion} utilize spatial and temporal guides. T2I-based extensions include AnimateDiff \cite{guo2023animatediff}, ModelScope \cite{wang2023modelscope}, and InstructVideo \cite{yuan2024instructvideo}.

\subsection{Video Motion Editing and Transfer}
Video motion control research has developed along two primary paths: explicit control through bounding boxes and motion transfer from reference videos. Explicit control methods include AnimateAnyone \cite{li2024animate}, Boximator \cite{wang2024boximator}, Peekaboo \cite{jain2024peekaboo}, and Trailblazer \cite{ma2023trailblazer}. 

Another significant line of work focuses on transferring motion from reference videos. MotionDirector (MD) \cite{zhao2025motiondirector} made a significant advancement with its innovative dual-path LoRA architecture, effectively separating motion and appearance characteristics through specialized components that enable precise control over temporal dynamics. DreamVideo \cite{wei2024dreamvideo} and Customize-A-Video \cite{ren2024customize} further refined this separation using distinct branches for appearance and motion learning. Motion Inversion \cite{wang2024motion} introduced a novel approach by learning motion embeddings through temporal attention layers trained directly on the original video.

 Video Motion Customization (VMC) \cite{jeong2024vmc} introduced a novel approach combining fine-tuning with inversion through adaptive temporal layer adjustments, achieving superior motion transfer results while maintaining temporal consistency. TokenFlow \cite{geyer2023tokenflow}, ReRender-A-Video \cite{yang2023rerender}, and RAVE \cite{kara2024rave} explored various approaches to temporal consistency. The field has further advanced with MotionInversion (MI) \cite{wang2024motion} that enable precise control over temporal dynamics while maintaining visual quality through sophisticated motion embeddings trained from a reference video.

A persistent challenge in motion transfer is the assumption of feature similarity between reference and target videos. DMT \cite{yatim2024space} addresses this limitation through a novel space-time feature loss, leveraging internal UNet activations for improved motion fidelity. This approach achieves superior results in maintaining temporal consistency while allowing for more diverse edited outputs compared to traditional feature-matching methods.

\begin{figure*}[t]
    \centering
    \begin{tabular}{c}
        \includegraphics[width=\textwidth]{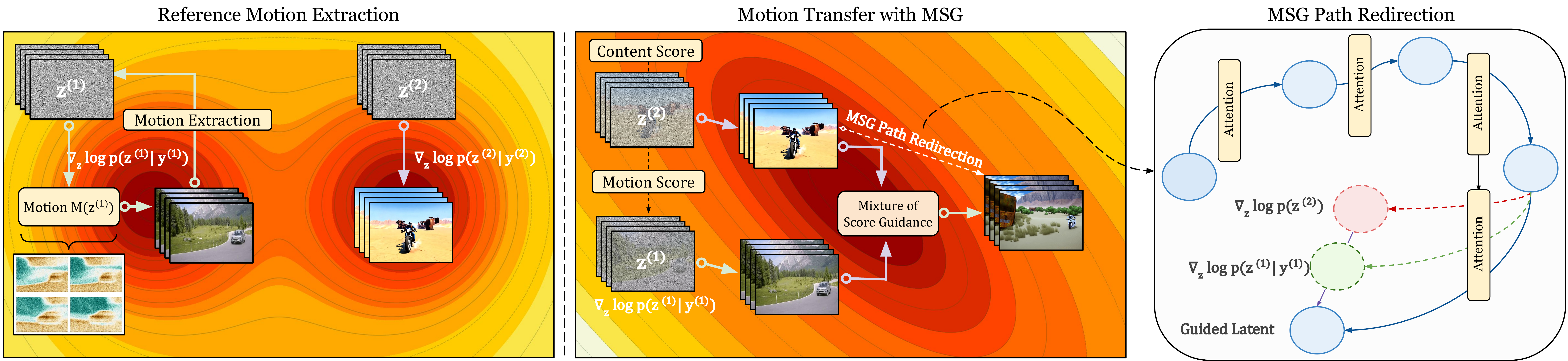} \\
    \end{tabular}
     \caption{\textbf{Method Overview.} Framework of our Mixture of Score Guidance (MSG) for zero-shot motion transfer in diffusion models. \textbf{Left:} Reference motion extraction stage captures motion characteristics $M(z)$ from early-timestep conditional scores $\nabla_z \log p(z^{(1)}|y^{(1)})$ and $\nabla_z \log p(z^{(2)}|y^{(2)})$. \textbf{Middle:} Motion transfer combines content and motion scores through our MSG formulation $s_{\text{MSG}}(z_t, z_t^*) = \nabla_z \log p_t(z|y) + w_{\text{MSG}}(\nabla_z \log p_t(z^*|y^*) - \nabla_z \log p_t(z))$. \textbf{Right:} MSG path redirection mechanism showing attention-guided dynamics that enable stable motion transfer by exploring the correct motion manifold while preserving content through modified Langevin dynamics governed by our mixture of potential energies $U_{\text{MSG}}(z_t) = U_{\text{content}}(z_t) + w_{\text{MSG}}[U_{\text{motion}}(z_t, z_t^*) - U_{\text{prior}}(z_t)]$.}
    \label{fig:framework}
\end{figure*}

\section{Background}
\textbf{Diffusion Process.} Consider a video sequence as a high-dimensional random variable $z \in \mathcal{Z}$ following an unknown data distribution $p(z)$. The diffusion process gradually transforms this distribution to a known prior distribution through a forward process defined by the following stochastic differential equation:
\begin{equation}
    dz = [f(z,t) - \frac{g(t)^2}{2}\nabla_z \log p_t(z)]dt + g(t)d\bar{w_t}
\end{equation}
where the drift coefficient $f(z, t)$ is characterized by:
\begin{equation}
    f(z, t) = -\dot{\sigma}(t)\sigma(t)\nabla_{z_t}\log p_t(z)dt
\end{equation}
and the diffusion coefficient $g(t)$ takes the form:
\begin{equation}
    g(t) = \sigma(t)\sqrt{2\beta(t)}
\end{equation}
The stochastic process is driven by the standard Wiener process $d\bar{w_t}$, while $p_t(z_t)$ represents the probability distribution of the noisy samples at time $t$. The boundary conditions of this distribution are given by the data distribution at the initial time, $p_0(z_0) = p_\text{data}(z)$, and a normal distribution with specified variance at the terminal time, $p_1(z_1) = \mathcal{N}(0, \sigma^2_\text{max}\mathbf{I})$. The time-reversed stochastic process for variance-preserving (VP) conditional diffusion is formulated in \cite{song2020score} as:
\begin{equation} \label{eq:reverse_SDE}
    dz = -\frac{1}{2}\beta_tzdt - \beta_t\nabla_z\log p_t(z|y)dt + \sqrt{\beta_t}\bar{w_t}
\end{equation}
By indicating directions of increased probability, the score naturally serves as a mechanism to undo the forward diffusion process. \\

\noindent \textbf{Classifier Free Guidance.} Classifier-Free Guidance (CFG) \cite{ho2022classifier} enhances generation quality by interpolating between conditional and unconditional score predictions, effectively balancing fidelity and diversity in the output.  CFG introduces a guided score $\nabla_z\log p_{t, \lambda}(z|y)$ that replaces the conditional score $\nabla_z\log p_t(z|y)$ in (\ref{eq:reverse_SDE}), defined at each timestep as:
\begin{equation} \label{eq:CFG}
    \nabla_z\log p_{t, \lambda}(z|y) = (1 - \lambda)\nabla_z\log p_t(z) + \lambda\nabla_z\log p_t(z|y)
\end{equation}
where $\lambda = 1$ reduces to the standard conditional generation, while $\lambda > 1$ amplifies the influence of the conditioning signal, typically leading to higher-quality but potentially less diverse outputs. \\

\noindent \textbf{Langevin Dynamics.} 
As a fundamental stochastic process in statistical physics, Langevin dynamics (LD)~\cite{parisi1981correlation, rossky1978brownian} enables sampling from complex probability distributions through continuous-time evolution. The dynamics follow a stochastic differential equation that combines deterministic drift with random fluctuations~\cite{robert1999monte}:
\begin{equation}
    dz = \frac{\epsilon}{2} \nabla\log p(z)dt + \sqrt{\epsilon}d\bar{w_t}
\end{equation}
When implemented with appropriate step sizes, this process naturally evolves toward its equilibrium state $p(z)$~\cite{roberts1996exponential}, making it particularly valuable for sampling tasks. The method's practical implementation hinges on the availability of the score function $\nabla\log p(z)$, which, similar to diffusion models, can be estimated through neural networks.
\section{Methodology}
\label{sec:methodology}
This section presents the theoretical foundations and formulations of Mixture of Score Guidance (MSG), a novel approach for motion transfer in diffusion models in terms of statistical mechanics and stochastic processes.

\subsection{Score-Based Motion Transfer}

\subsubsection{Score Function Decomposition}
Let $\mathcal{M}: \mathbf{Z} \rightarrow \mathbf{M}$ be a mapping from the latent space to motion characteristics. The score function $\nabla_z \log p_t(z|y)$ can be separated into motion and content components through our conditional reformulation $\nabla_{z} \log p_t(z, \mathcal{M}(z^*)|y)$:
\begin{align*}
    \nabla_{z} \log p_t(z, \mathcal{M}(z^*)|y) &= \nabla_{z} \log p_t(\mathcal{M}(z^*)|y)  \\
                                                                &+ \nabla_{z} \log p_t(z|\mathcal{M}(z^*),y)
\end{align*}
where $\mathcal{M}(z^*)$ is a reference motion representation and it is a function of the reference video latent $z^*$. This decomposition separates the score function into two meaningful components: 

\noindent\textbf{(1) Motion Score:} $\nabla_{z} \log p_t(\mathcal{M}(z^*)|y)$ which is responsible for capturing how the latent affects motion characteristics and representing the gradient of log-likelihood concerning motion. The term dominates in early timesteps due to motion's hierarchical nature. 

\noindent \textbf{(2) Content Score:} $\nabla_{z} \log p_t(z|\mathcal{M}(z^*),y)$ which captures content-specific information conditioned on motion and represents the residual gradient after accounting for motion. As a result it is more prominent in later timesteps.

\subsubsection{Mixture of Score Guidance}
Given a reference video with desired motion characteristics characterized by the reference condition $y^*$ with the latents $z^*$, we formulate MSG as:
\vspace{-5pt}
\begin{align*}
    s_\text{MSG}(z_t, z^*_t) &= \nabla_{z} \log p_t(z|y) \\
    &+w_{\texttt{MSG}}(\nabla_{z} \log p_t(z^*|y^*) - \nabla_{z} \log p_t(z))
\end{align*}

This formulation can be interpreted as a statistical mixture model in score space, where each component contributes to different aspects of the generation process. The theoretical significance of MSG can be understood through its relationship with Langevin dynamics. Consider the standard  Langevin equation:
\begin{equation}
    dz_t = \nabla_z U(z_t)dt + \sqrt{2\beta^{-1}}dW_t
\end{equation}
where $U(z_t)$ is the potential energy function and $\beta$ is the inverse temperature. Our MSG formulation extends this to a mixture of potential energies:
\begin{equation}
    U_\text{MSG}(z_t) = U_\text{content}(z_t) + w_{\texttt{MSG}}[U_\text{motion}(z_t, z^*_t) - U_\text{prior}(z_t)]
\end{equation}
This leads to the modified Langevin dynamics:
\begin{equation}
    dz_t = \nabla_z U_\text{MSG}(z_t)dt + \sqrt{2\beta^{-1}}dW_t
\end{equation}
Since the proposed operation does not harm the original dynamics of the denoising process, the system explores the correct motion manifold while preserving content and the resulting trajectories are stable and free from spurious artifacts.
\begin{figure*}[t!]
    \centering
    \begin{tabular}{c}
        \includegraphics[width=\textwidth]{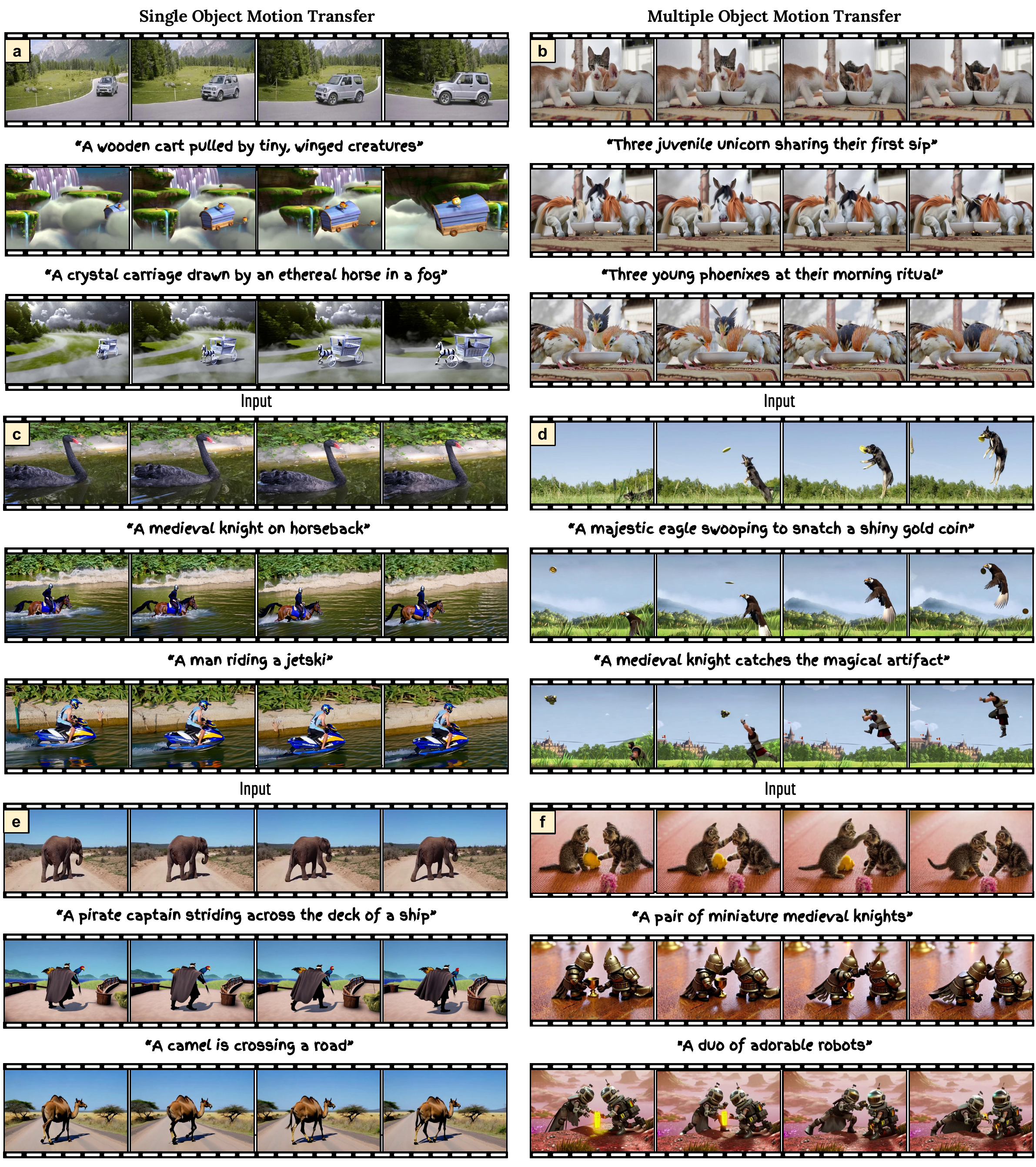} \\
    \end{tabular}
     \caption{Qualitative results demonstrating our method's ability to preserve motion priors while generating novel content from text prompts. \textbf{(Left)} Single-object motion transfer where complex motions like mechanical movements, horseback riding sequences are accurately preserved in the generated outputs. \textbf{(Right)} Multi-object scenarios where our method successfully maintains the original motion dynamics while generating diverse subjects. Please refer to the Supplementary Material for full videos and additional examples. }
    \label{fig:qual}
\end{figure*}
\subsubsection{Motion Trajectory Representation}
\noindent Let $\mathcal{V} \in \mathbb{R}^{F \times H \times W \times 3}$ denote an input video sequence. Our aim is to derive a training-free motion representation through the following formulation. The forward process transforms the initial frame latents of the reference $z^*_0$ into noised latents $z^*_t$ at timestep $t$ according to:
\begin{equation}
    z^*_t = \alpha_t z^*_0 + \sigma_t \epsilon, \quad \epsilon \sim \mathcal{N}(0, I)
\end{equation}
where $\alpha_t$ and $\sigma_t$ are time-dependent coefficients controlling the noise schedule \cite{ho2020denoising}. The noising step is controlled by the strength parameter. The conditional score function $\nabla_z\log p_t(z^*|y)$ is then computed via a pretrained denoising network. Through investigation of score distribution, we establish that the motion representation operator $\mathcal{M}: \mathcal{Z} \rightarrow \mathcal{Z}$ defined as $\mathcal{M}(z) = \nabla_{z_t} \log p_t(z|y)$ 
captures predominant motion patterns at early diffusion timesteps $t \ll T$. This finding is empirically validated through our analysis of reference video conditional noise patterns, as illustrated in Figure~\ref{fig:motion_representation} and Figure~\ref{fig:framework}.

 \section{Experiments}
\label{sec:experiments}

\begin{figure*}[t]
    \centering
    \begin{tabular}{c}
        \includegraphics[width=\linewidth]{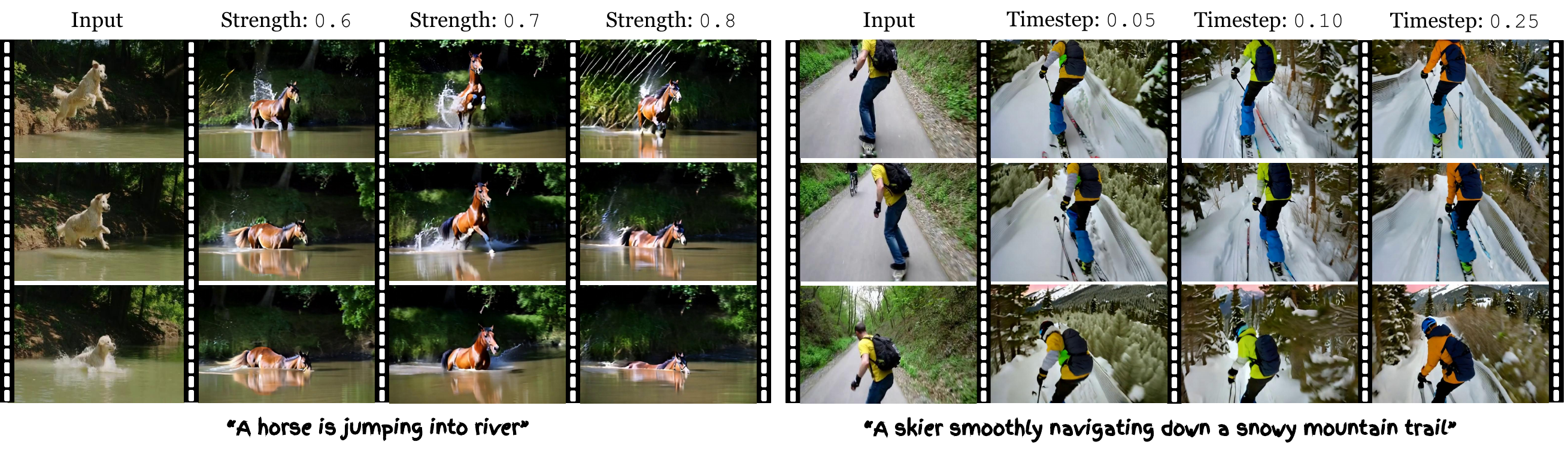} \\
    \end{tabular}
     \caption{\textbf{Ablation study on strength and timestep parameters.} Left: We analyze the effect of noise addition in the motion extraction stage, where strength=0.7 achieves optimal motion representation - lower values (0.6) result in weak motion transfer while higher values (0.8) lead to over-stylization. Right: Impact of applying Mixture of Score guidance at different timestep ratios of total 50 timesteps on motion transfer quality.}
    \label{fig:ablation_guidance}
\end{figure*}

\noindent \textbf{Experimental Setup.} Our implementation utilizes the CogVideoX \cite{yang2024cogvideox} model for video generation and editing. We conduct all experiments at a resolution of 720 x 480 pixels using 50 diffusion timesteps. Due to the absence of a dedicated DDIM inversion schedule in CogVideoX, we employ a stochastic inversion approach where we add controlled noise to the input video latents, regulated by a strength parameter (detailed analysis in Fig.~\ref{fig:ablation_guidance}). For motion transfer, our pipeline operates in two phases: first, we obtain conditional score estimates from the reference video in early timesteps ($t \ll T$), then we apply this guidance during the generation of motion-transferred videos at the same timestep range. Throughout all experiments presented in this paper, we consistently set t to 10\% of the total timesteps, as this configuration provides an effective balance between motion preservation and generation quality. 

\section{Qualitative Experiments}
\begin{figure*}[t]
    \centering
    \begin{tabular}{c}
        \includegraphics[width=\textwidth]{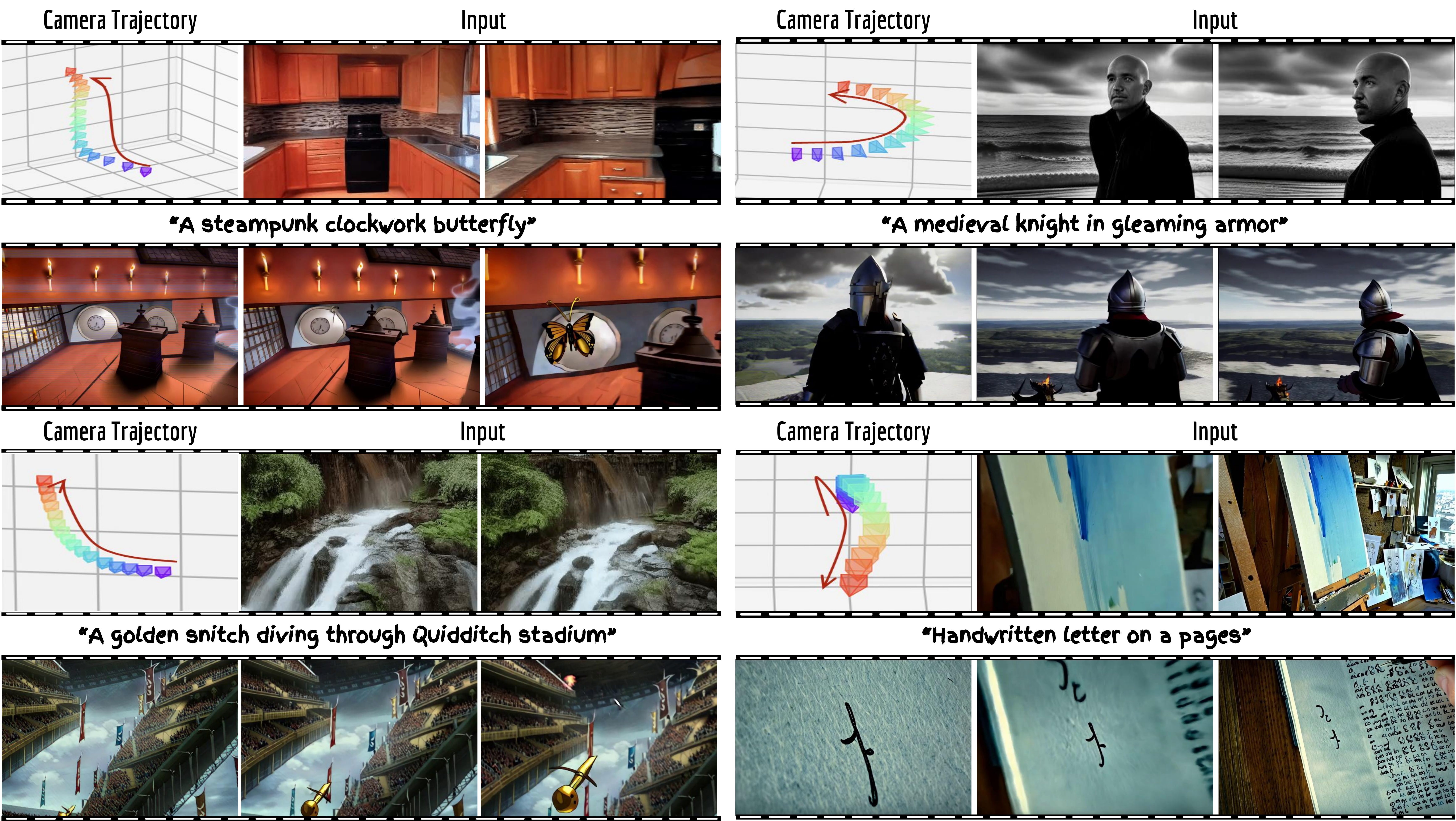} \\
    \end{tabular}
     \caption{\textbf{Camera Motion Transfer Results Across Diverse Scenarios.} Each row shows the camera trajectory (left) and corresponding input-output image sequences. Our method can transfer camera motions while maintaining spatial consistency, as demonstrated in various cases: a steampunk clockwork butterfly animation, a raindrop on a leaf, an eagle soaring through mountain peaks, and dominos falling on a rail track. The colored trajectories represent the camera path through 3D space, with different colors indicating temporal progression.}
\label{fig:camera_motion_transfer}
\end{figure*}

Our experimental results demonstrate MotionShop's versatility across diverse motion transfer scenarios. As shown in Figures \ref{fig:teaser} and \ref{fig:qual}, our method successfully handles both single and multi-object transfers. For single-object scenarios, MotionShop effectively transforms a black swan into a horse and a man riding jet ski (Fig. \ref{fig:qual}.c), maintaining realistic movement patterns and contextual elements like water splashes. In multi-object cases, our method seamlessly converts cats into birds (Fig. \ref{fig:qual}.b) and robots (Fig. \ref{fig:qual}.f). MotionShop provides flexible background control—enabling both dramatic alterations (Fig. \ref{fig:qual}.a) and preservation according to the text prompt (Fig. \ref{fig:qual}.c). Our method also supports concurrent motion controls, demonstrated by simultaneously transforming a frisbee into a coin while converting a dog into an eagle (Fig. \ref{fig:qual}.d). Additionally, MotionShop handles complex camera movements including zoom-ins, zoom-outs (Fig. \ref{fig:teaser}), and rotational movements (Fig. \ref{fig:camera_motion_transfer}), showcasing its comprehensive motion transfer capabilities.

\begin{figure*}[t]
    \centering
    \begin{tabular}{c}
        \includegraphics[width=\linewidth]{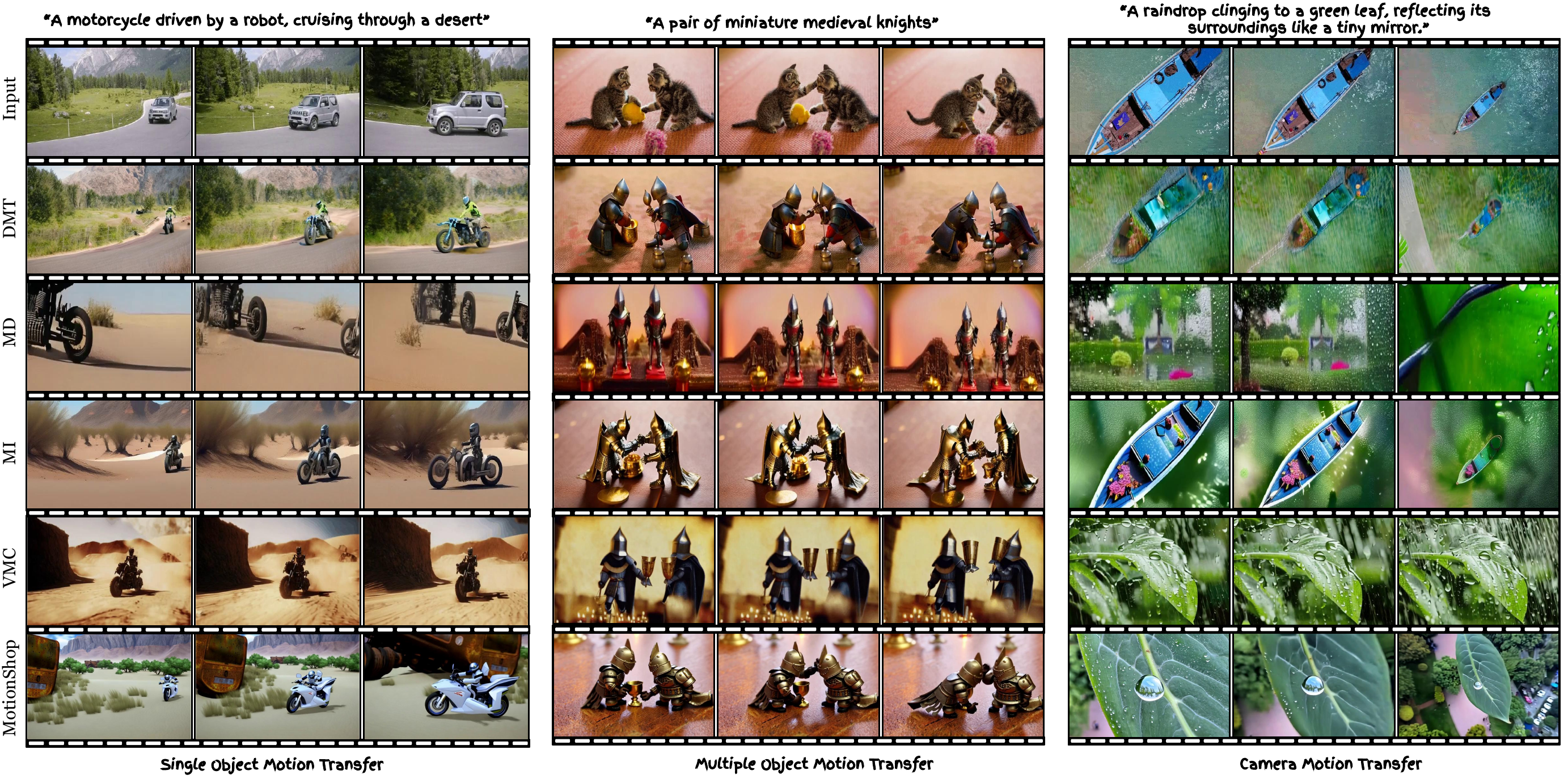} \\
    \end{tabular}
    \caption{\textbf{Qualitative comparison of motion transfer capabilities.} We compare \texttt{MotionShop} (bottom row) with existing methods (VMC, DMT, MD, MI) on three challenging scenarios. Left: Single object motion transfer of a robot-driven motorcycle in a desert scene. Middle: Multiple object motion transfer involving miniature medieval knights, demonstrating the ability to preserve interactions between objects. Right: Camera motion transfer capturing the dynamic perspective of a raindrop on a leaf. Our method demonstrates superior motion-text alignment across all three motion transfer categories.}
\label{fig:qualitative_comparison}
\end{figure*}

\begin{table*}[t]
\centering
\definecolor{lightgreen}{RGB}{240,255,240}
\begin{tabular}{l|ccc|c|ccc}
\toprule
\multirow{2}{*}{Method} & \multicolumn{3}{c|}{Quantitative Metrics} & \multicolumn{4}{c}{User Study} \\
\cmidrule(lr){2-4} \cmidrule(lr){5-8}
 & Text Sim.↑ & Motion Fid.↑ & Temp. Cons.↑ & FID↓ & Text Sim.↑ & Motion Fid.↑ & Temp. Cons.↑ \\
\midrule
DMT~\cite{yatim2024space} & 0.298 & 0.884 & 0.911 & \textbf{196.54} & 0.21 & 0.19 & 0.20 \\
VMC~\cite{jeong2024vmc} & \textbf{0.328} & 0.380 & 0.924 & 237.15 & 0.06 & 0.06 & 0.15 \\
MD~\cite{zhao2025motiondirector} & 0.285 & 0.828 & 0.904 & 222.92 & 0.13 & 0.14 & 0.10 \\
MI~\cite{wang2024motion} & 0.304 & 0.735 & 0.735 & 210.90 & 0.19 & 0.18 & 0.17 \\
\midrule
\rowcolor{lightgreen}
Ours & 0.314 & \textbf{0.913} & \textbf{0.928} & 209.06 & \textbf{0.41} & \textbf{0.43} & \textbf{0.38} \\
\bottomrule
\end{tabular}
\caption{\textbf{Comprehensive Analysis of Motion Generation Methods.} We evaluate our approach against state-of-the-art methods using both quantitative metrics (Text Similarity, Motion Fidelity, Temporal Consistency, and FID) and human evaluation. Our method achieves superior performance in most metrics, particularly showing significant improvements in user studies. Arrows (↑/↓) indicate higher/lower values are better, and best results are shown in \textbf{bold}.}
\label{tab:comparison}
\end{table*}

\begin{figure}[h!]
    \centering
    \begin{tabular}{c}
    \hspace{-0.3cm}
        \includegraphics[width=1\linewidth]{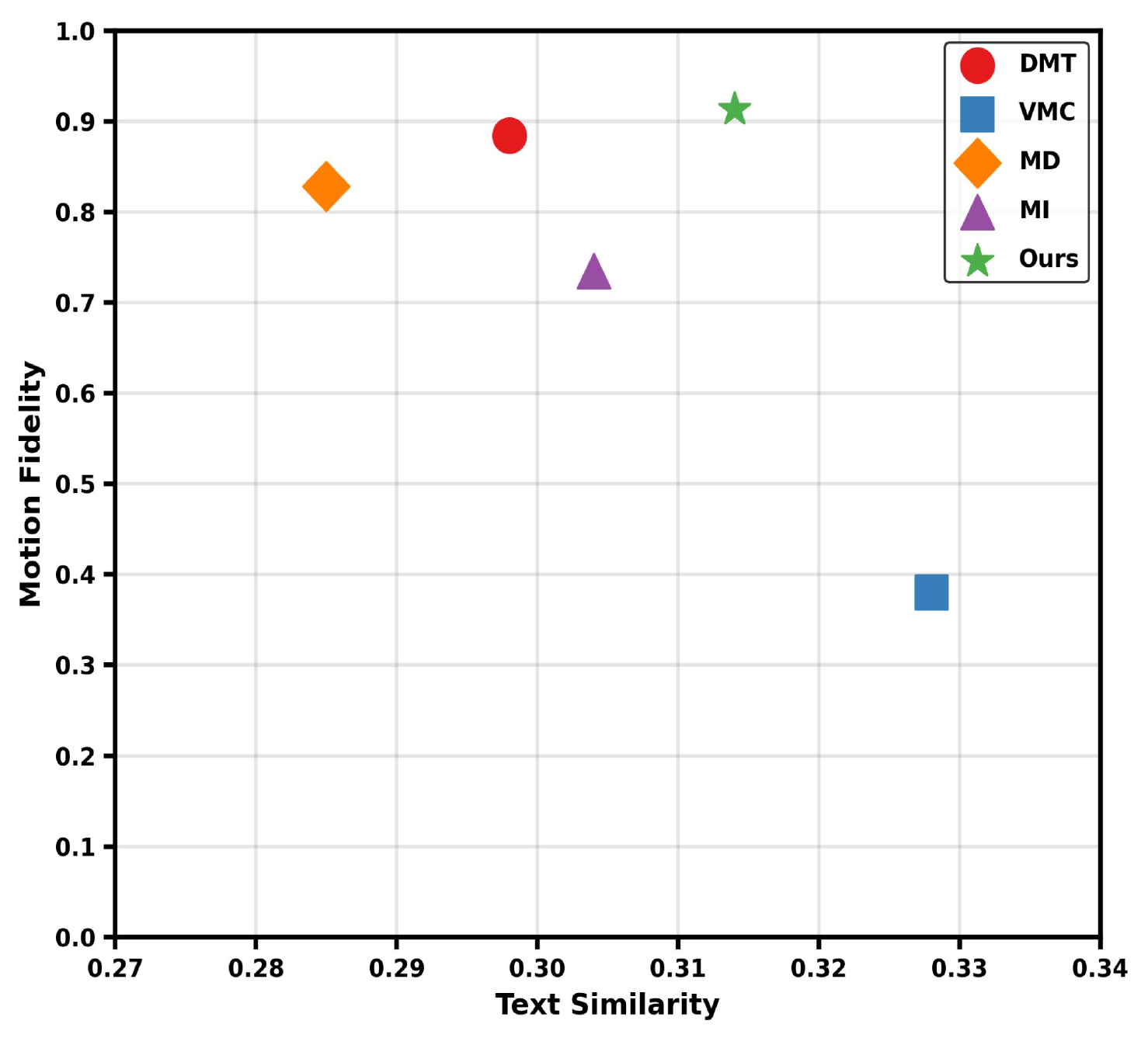} \\
    \end{tabular}
    \caption{\textbf{Trade-off Analysis between Text Similarity and Motion Fidelity.} Comparison of our method against baselines shows superior performance in both metrics, with our approach (green star) achieving higher motion fidelity (0.913) while maintaining competitive text similarity (0.314).}
    \label{fig:tradeoff}
\end{figure}

\begin{figure*}[h!]
    \centering
    \begin{tabular}{c}
    \hspace{-0.3cm}
        \includegraphics[width=1\linewidth]{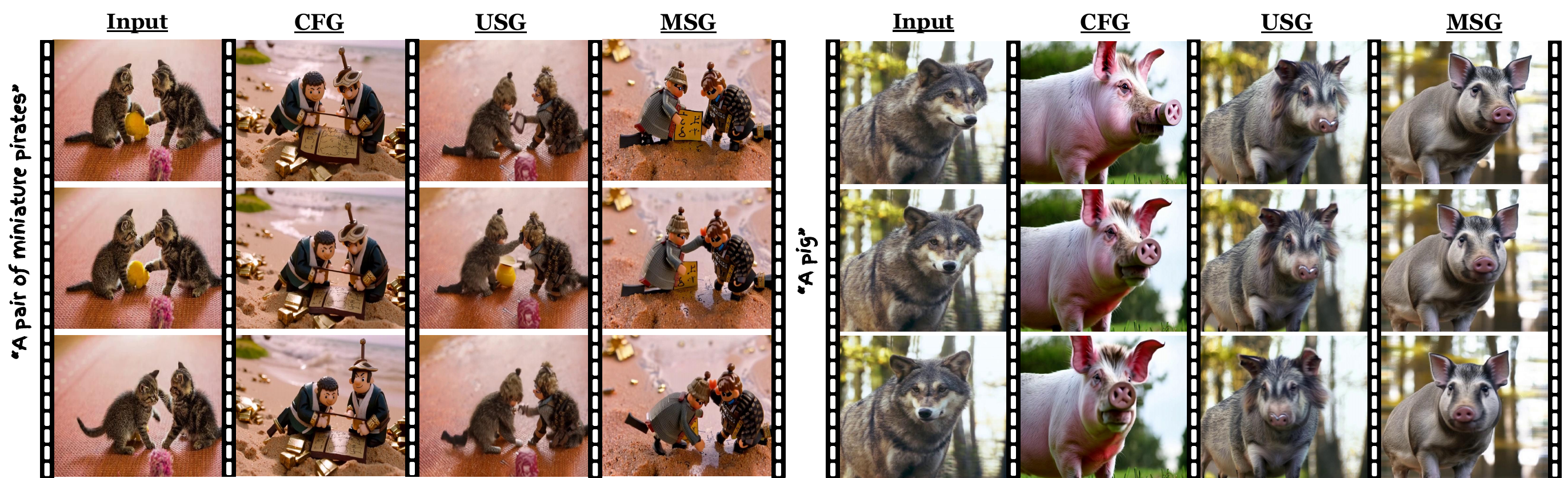} \\
    \end{tabular}
    \caption{\textbf{Comparison of different guidance mechanisms.} Comparing our Mixture of Score Guidance (MSG) against Classifier-Free Guidance (CFG, baseline without reference) and Unconditional Score Guidance (USG, using reference video's unconditional score).}
    \label{fig:ablation_score}
\end{figure*}

\section{Qualitative Comparisons}
 We conducted a qualitative comparison of MotionShop against MotionInversion \cite{wang2024motion}, DMT \cite{yatim2024space}, VMC \cite{jeong2024vmc}, and MotionDirector \cite{zhao2025motiondirector}, as shown in Table~\ref{fig:qualitative_comparison}. Our evaluation focused on motion transfer across single/multiple objects and complex camera movements. Our experimental results reveal key differences in background handling among the methods: MotionDirector and VMC show limitations in background preservation, introducing undesirable artifacts (Table~\ref{fig:qualitative_comparison}). In contrast, MotionShop demonstrates two distinct advantages: it enables accurate background modification when explicitly requested in the prompt (Table~\ref{fig:qualitative_comparison}), and maintains consistent preservation of the original scene composition while properly transforming target objects (Table~\ref{fig:qualitative_comparison}). These results indicate MotionShop's superior ability to distinguish between intentional and unintentional scene modifications. Additionally, MotionShop excels in transferring complex camera movements, including zoom-ins, zoom-outs, and rotations and their combinations such as pan-left and zoom-out (Fig.~\ref{fig:camera_motion_transfer} bottom right), which proved challenging for DMT and MotionDirector in creative scene camera motion transfer (Fig.~\ref{fig:tradeoff}).

\section{Quantitative Experiments}
In our quantitative evaluation, we compared MotionShop with MotionInversion \cite{wang2024motion}, DMT \cite{yatim2024space}, VMC \cite{jeong2024vmc}, and MotionDirector \cite{zhao2025motiondirector} with 100 data-prompt pairs using four metrics: (1) \textit{Text Similarity}, measuring frame-to-text alignment using CLIP \cite{radford2021learning}, (2) \textit{Motion Fidelity} \cite{yatim2024space}, evaluating motion preservation using tracklet similarity between input and output videos, (3) \textit{Temporal Consistency}, measuring frame-to-frame coherence via CLIP feature similarity, and (4) \textit{FID}, assessing visual quality against DAVIS dataset. As shown in Table~\ref{tab:comparison}, MotionShop achieves state-of-the-art performance in both \textit{Motion Fidelity} (0.913) and \textit{Temporal Consistency} (0.928). While VMC shows marginally higher \textit{Text Similarity} (0.328 vs. 0.314), our method achieves a better balance between text alignment and motion quality metrics (Table~\ref{tab:comparison}).

\section{Discussion on Quantitative Experiments}
The quantitative evaluation results presented in Table~\ref{tab:comparison} demonstrate the superior performance of our approach. Specifically, MotionShop achieves state-of-the-art performance in Motion Fidelity, surpassing the previous best method (DMT~\cite{yatim2024space}) by a significant margin of 2.9\%. In terms of Text Similarity metrics, our quantitative analysis reveals an interesting trade-off between text alignment and motion preservation. While VMC~\cite{jeong2024vmc} achieves marginally higher Text Similarity scores (surpassing MotionShop by 1.4\%), our experimental results indicate that this advantage comes at a significant cost to motion fidelity. In contrast, MotionShop maintains competitive text alignment capabilities (second-best among all methods) while simultaneously achieving superior motion preservation, demonstrating a more balanced approach to the inherent text-motion trade-off in video generation tasks. Analysis of the Fréchet Inception Distance (FID) reveals that our method achieves competitive performance, ranking second with a margin of \textit{12.52} compared to DMT~\cite{yatim2024space}. However, deeper examination of both Text-Similarity metrics and broader quantitative results provides critical context for these FID scores. While DMT exhibits lower FID values, this advantage appears to stem from its conservative approach to scene modification, predominantly preserving original scene layouts and compositions rather than implementing creative transformations as specified in the prompts. This characteristic leads to numerically favorable FID scores but potentially limits the method's utility for more ambitious motion transfer applications requiring significant scene modifications. Our approach, in contrast, demonstrates a more balanced capability, successfully executing substantial scene transformations while maintaining reasonable FID scores, thus offering greater practical utility for diverse motion transfer scenarios.

\noindent \textbf{User Study.} We conducted a user study with $N=50$ participants on Prolific.com, evaluating 30 sets of videos. Participants assessed three metrics by selecting the top two results for each: Motion Preservation, Temporal Consistency, and Text Alignment. Results in Table~\ref{tab:comparison} demonstrate MotionShop's consistent superiority across all metrics, outperforming existing approaches in motion preservation, temporal coherence, and text-guided modifications (see Appendix for more details).

\subsection{Ablation Studies}
We analyze three key components: motion extraction strength, guidance timestep ratio, and guidance mechanisms. Fig.~\ref{fig:ablation_guidance} (left) shows the impact of motion extraction strength parameters. At 0.6, the horse's jumping motion is insufficiently transferred; at 0.8, over-stylization distorts the motion; 0.7 achieves optimal balance between motion preservation and visual quality. Fig.~\ref{fig:ablation_guidance} (right) demonstrates that applying Mixture of Score guidance at different timesteps preserves generative priors, enabling diverse yet natural jumping motions.

Fig.~\ref{fig:ablation_score} compares three guidance approaches: Classifier-Free Guidance (CFG), Unconditional Score Guidance (USG), and our Mixture of Score Guidance (MSG). CFG struggles with motion consistency, while USG better preserves motion but lacks prompt-guided precision. MSG demonstrates superior performance, evidenced by natural medieval cat motions (left) and wolf-to-pig transformations (right) while maintaining motion characteristics. This improvement stems from our novel formulation that explicitly decomposes motion and content scores, allowing for more precise control over the transfer process. 

\section{MotionBench Dataset}
We introduce MotionBench, a comprehensive motion transfer dataset designed for systematic evaluation of motion transfer capabilities. The dataset comprises 200 carefully curated source videos and 1,000 corresponding transferred sequences, combining real-world footage from DAVIS dataset (50 videos) and high-quality synthetic videos (150 videos) generated using CogVideoX [33].

The dataset is structured around three primary motion categories, each addressing distinct challenges in motion transfer: Single Object Motion (85 videos, 42.5\%), Multiple Object Motion (65 videos, 32.5\%), and Camera Motion (50 videos, 25\%). Single object sequences capture diverse motion patterns from rigid mechanical movements to complex articulated motions. Multiple object scenarios evaluate preservation of spatial relationships and interaction dynamics between moving entities. Camera motion sequences test the handling of viewpoint changes through both simple camera operations (pan, tilt, zoom) and complex trajectories combining multiple movement types.

Each source video is paired with multiple target motion transfers, systematically exploring scenarios from straightforward object-to-object transformations to comprehensive scene-level modifications. The transfers evaluate both motion fidelity and creative adaptation capabilities, ranging from preserving precise mechanical movements to transferring organic motion patterns onto radically different targets. For example, the dataset includes challenging cases like transferring vehicle motion to flying creatures while preserving trajectory dynamics. All videos maintain consistent technical specifications (720$\times$480 resolution) to enable standardized evaluation. We provide dataset statistics, category descriptions, and comprehensive analysis of motion transfer scenarios in the supplementary material.

\section{Limitation and Societal Impact.}
\label{sec:conclusion}
Our method's performance is inherently tied to the generative priors learned by the underlying T2V model. As a result, certain target concepts and motions may fall outside the model's distribution. Additionally, any biases present in the T2V model are carried over into our approach, a drawback that any zero-shot method suffers, which may influence the quality of generated outputs for specific scenarios. Since our method enables controllable video generation, there is a potential risk of it being used to create deepfake videos that spread misinformation or deceive viewers. To mitigate these risks, we emphasize the importance of ethical use of our tool.\\
\section{Conclusion}
In this paper, we presented MotionShop, the first motion transfer approach in video diffusion transformers, which reformulates conditional score to decompose motion and content scores. By treating motion transfer as a mixture of potential energies, our method enables creative scene transformations while preserving motion patterns, operating directly on pre-trained models without additional training. Extensive experiments demonstrate MSG's effectiveness across various scenarios, from single/multiple object transformations to complex camera motion transfer. Our framework provides principled guidance for balancing motion and content preservation, enabling flexible motion transfer in video generation.

{
    \small
    \bibliographystyle{ieeenat_fullname}
    \bibliography{main}
}

\clearpage
\setcounter{page}{1}
\setcounter{section}{0}  %
\renewcommand{\thesection}{\Alph{section}}  %
\maketitlesupplementary
\begin{figure*}[t]
    \centering
    \begin{tabular}{c}
        \includegraphics[width=\linewidth]{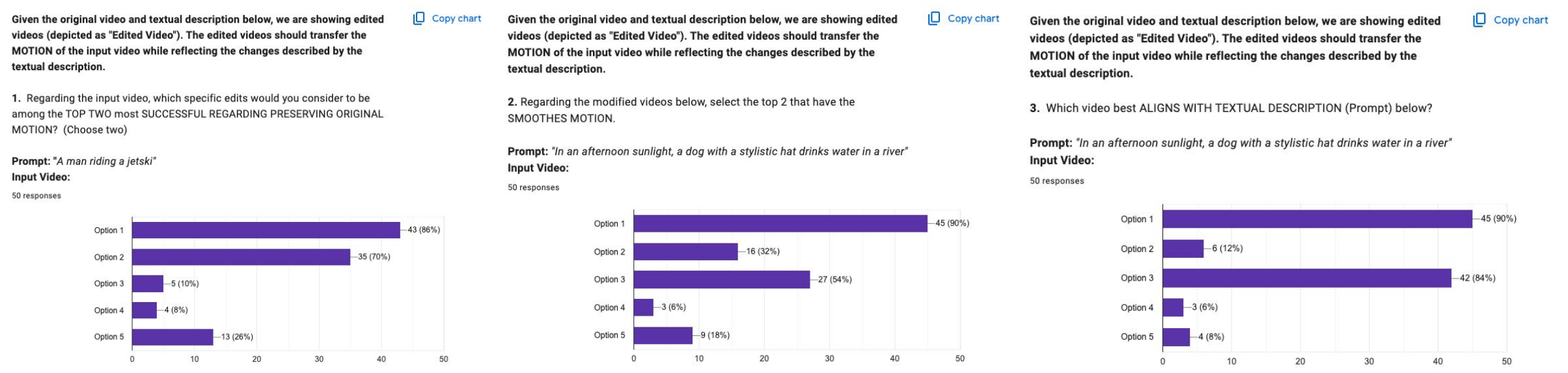} \\
    \end{tabular}
    \caption{\textbf{Type of Questions.} We ask 3 different questions for Text Alignment, Motion Fidelity and Temporal Consistency.}
\label{fig:user_study_details}
\end{figure*}

\section{User Study Details}
\label{sec:user_study_details}

To evaluate the perceptual quality of our method, we conducted a comprehensive user study with N=50 participants recruited through Prolific.com. Following standard practices in human evaluation studies for video generation \cite{wang2024motion}, we designed our study to assess three critical aspects of motion transfer quality, as illustrated in Fig.~\ref{fig:user_study_details}.

For each test case, participants were presented with an input video and five different edited versions, corresponding to various motion transfer methods. The evaluation criteria were as follows:

\begin{enumerate}
    \item \textbf{Motion Fidelity:} Participants were asked to identify the two edited videos that best preserved the motion patterns from the input sequence. This assessment focused on the accuracy of transferred motion dynamics and spatial relationships. The question we asked is being \textit{"Regarding the input video, which specific edits would you consider to be among the top two most successful regarding preserving original motion?"}
    
    \item \textbf{Temporal Consistency:} Users selected the two results exhibiting the highest temporal coherence, evaluating frame-to-frame continuity and the absence of artifacts or jitter in the generated sequences. The question we asked is being \textit{"Regarding the modified videos below, select the top 2 that have the smoothest motion."}
    
    \item \textbf{Text-Motion Alignment:} Participants evaluated how well each generated video aligned with its corresponding text prompt, focusing on both semantic accuracy and motion appropriateness. The question we asked is being \textit{"Which video best aligns with textual description (prompt) below."}
\end{enumerate}

The study compared five different approaches: our proposed MotionShop method, Space-Time Features (DMT)~\cite{yatim2024space}, MotionDirector (MD)~\cite{zhao2025motiondirector}, MotionInversion (MI)~\cite{wang2024motion}, and Video Motion Customization (VMC)~\cite{jeong2024vmc}. The user study interface, shown in Fig.~\ref{fig:user_study_interface}, was designed to facilitate clear comparison and intuitive interaction.

\section{MotionBench: A Comprehensive Motion Transfer Dataset} 
\label{sec:motionbench}

We introduce MotionBench, the first publicly available dataset specifically designed for evaluating motion transfer capabilities in video generation models. While existing video datasets primarily focus on general video synthesis or editing tasks, MotionBench addresses the critical gap in standardized evaluation of motion transfer capabilities. The dataset comprises 200 carefully curated source videos and 1,000 corresponding motion-transferred sequences, enabling systematic evaluation across diverse motion patterns and scene compositions.

\subsection{Dataset Composition}
\label{subsec:composition}

\subsubsection{Source Videos}
The 200 source videos are curated from two primary sources:
\begin{itemize}
    \item DAVIS Dataset (50 videos): Selected for their diverse real-world motions
    \item Synthetic Videos (150 videos): Generated using CogVideoX-5B \cite{yang2024cogvideox} model.
\end{itemize}

\noindent The source videos are categorized into the following motion categories:

\noindent\textbf{1. Single Object Motion (85 videos)}

The Single Object Motion category constitutes the largest portion of our dataset (42.5\% of source videos), carefully curated to capture the full spectrum of motion patterns observed in real-world scenarios. This category is subdivided into three distinct motion types:

\noindent\textbf{Rigid Object Motion (35 videos):} This subcategory focuses on objects that maintain their shape during motion, featuring vehicles (e.g., cars, motorcycles), toys (e.g., remote-controlled cars, mechanical toys), and mechanical objects (e.g., robotic arms, industrial machinery). These sequences are particularly valuable for evaluating a method's ability to preserve consistent object geometry while transferring motion patterns. 

\noindent\textbf{Non-rigid Object Motion (30 videos):} This subset encompasses objects that undergo deformation during movement, primarily featuring animals (e.g., birds in flight, running quadrupeds) and deformable objects (e.g., cloth, fluid-like materials). These sequences present more complex challenges, requiring methods to handle both global motion and local deformations simultaneously. The videos capture various natural movements including galloping, flying, and elastic deformations.

\begin{table}[t]
\centering
\caption{Distribution of videos across different motion categories in MotionBench. The dataset provides a balanced representation of various motion types, enabling comprehensive evaluation of motion transfer methods.}
\label{tab:motion_distribution}
\begin{tabular}{@{}lcc@{}}
\toprule
\textbf{Motion Type} & \textbf{Source Videos} & \textbf{Transfer Sequences} \\
\midrule
Single Object & 85 (42.5\%) & 400 (40\%) \\
Multi-Object & 65 (32.5\%) & 300 (30\%) \\
Camera Motion & 50 (25\%) & 300 (30\%) \\
\bottomrule
\end{tabular}
\end{table}

\noindent\textbf{Human Motion (20 videos):} The human motion sequences capture a diverse range of articulated movements, including walking sequences, dance performances, and various sports activities. These videos are particularly challenging as they combine both rigid (skeletal) and non-rigid (soft tissue) motion patterns. The sequences test a method's capability to preserve complex kinematic chains and natural human dynamics while transferring motion to different target subjects or characters.

Each subcategory is carefully balanced to include both simple and complex motion patterns, varying speeds, and different environmental contexts. This structured approach enables systematic evaluation of motion transfer methods across a spectrum of complexity levels, from basic rigid transformations to highly articulated and deformable motion patterns.

\noindent\textbf{2. Multi-Object Motion (65 videos)}

The Multi-Object Motion category comprises 32.5\% of our dataset, specifically designed to evaluate motion transfer capabilities in scenarios involving multiple moving entities. This category presents unique challenges in preserving spatial relationships, temporal synchronization, and complex interaction patterns. We organize these sequences into three distinct subcategories:

\noindent\textbf{Interactive Motion (25 videos):} These sequences capture complex interactions between multiple objects or animals, such as predator-prey chase sequences, children playing with toys, or animals engaged in social behaviors. The defining characteristic of this subset is the causal relationship between the subjects' movements, where the motion of one entity directly influences others. These videos are particularly challenging for motion transfer as they require preserving not only individual motion patterns but also the intricate timing and spatial relationships that define the interactions. Examples include dogs playing with frisbees, cats interacting with toys, and people passing objects between them.

\noindent\textbf{Independent Motion (20 videos):} This subcategory features scenarios where multiple objects move simultaneously but independently of each other. These sequences test a method's ability to maintain distinct motion patterns while ensuring global scene coherence. Examples include traffic scenes with multiple vehicles, scenes of birds flying in different directions, and sequences of independent mechanical systems operating simultaneously. The primary challenge lies in preserving the independence of various motion patterns while maintaining their temporal alignment and avoiding unintended interactions in the transferred results.

\noindent\textbf{Group Motion (20 videos):} The group motion sequences focus on coordinated movements of multiple subjects, such as synchronized dancing, flock behaviors, or team sports activities. These videos present unique challenges in maintaining both individual motion fidelity and group-level patterns. The sequences capture various forms of collective behavior, from highly structured (e.g., marching bands, synchronized swimming) to more organic patterns (e.g., school of fish, crowd movements). The key evaluation aspect is the preservation of both individual dynamics and emergent group behavior patterns during motion transfer.

Each subcategory is carefully curated to include varying levels of complexity in terms of the number of objects, spatial distribution, and temporal coordination. This structured organization enables comprehensive evaluation of how motion transfer methods handle scenarios ranging from simple multi-object scenes to complex, interdependent motion patterns, providing insights into their scalability and robustness in real-world applications.

\noindent\textbf{3. Camera Motion (50 videos)}

The Camera Motion category constitutes 25\% of our dataset, specifically designed to evaluate motion transfer methods' capabilities in handling various camera movement patterns. This category is particularly crucial as camera motion adds an additional layer of complexity to the motion transfer task, requiring methods to maintain coherent scene composition while adapting to changing viewpoints and perspectives.

\noindent\textbf{Simple Camera Movements (20 videos):} This subcategory encompasses fundamental camera operations that form the building blocks of cinematographic techniques. Each type presents unique challenges for motion transfer:

\begin{itemize}
    \item \textbf{Pan (5 videos):} Horizontal camera rotations that test a method's ability to maintain consistent object appearance and motion during lateral viewpoint changes. These sequences include landscape shots, architectural surveys, and subject tracking, with varying pan speeds and ranges.
    
    \item \textbf{Tilt (5 videos):} Vertical camera rotations that challenge perspective preservation, particularly in maintaining proper scale relationships as the viewing angle changes. Examples include vertical scans of buildings, waterfalls, and ascending/descending subject movements.
    
    \item \textbf{Zoom (5 videos):} Sequences involving camera focal length changes, testing a method's capability to handle continuous scale variations while preserving motion coherence. These include both zoom-in sequences revealing fine details and zoom-out shots revealing broader context.
    
    \item \textbf{Dolly (5 videos):} Forward/backward camera translations that evaluate depth handling and parallax effects. These shots are particularly challenging as they require maintaining proper spatial relationships between foreground and background elements during motion transfer.
\end{itemize}

\noindent\textbf{Complex Camera Movements (30 videos):} This subcategory features more sophisticated camera work that combines multiple basic movements, presenting higher-level challenges for motion transfer systems:

\begin{itemize}
    \item \textbf{Combined Motion Patterns (15 videos):} These sequences feature simultaneous execution of multiple camera movements (e.g., pan-with-zoom, tilt-with-dolly). They test a method's ability to handle compound camera transformations while maintaining scene coherence and motion fidelity. Examples include aerial shots with multiple degrees of freedom, elaborate reveal sequences, and complex establishing shots.
    
    \item \textbf{Dynamic Tracking Shots (15 videos):} These sequences involve camera movements that actively follow moving subjects, requiring simultaneous handling of both camera and subject motion patterns. They present particularly challenging scenarios where the camera movement must maintain a specific spatial relationship with the tracked subject while adapting to the subject's motion. Examples include sports coverage, chase sequences, and nature documentaries.
\end{itemize}

The camera motion sequences are carefully selected to include variations in speed, acceleration, and motion smoothness. Additionally, they encompass different environmental contexts (indoor/outdoor, varying lighting conditions) and subject types, providing a comprehensive evaluation framework for testing motion transfer methods' robustness to camera movement. This category is particularly valuable for assessing a method's potential in real-world applications such as cinematography, virtual production, and automated video editing.

\subsection{Motion Transfer Sequences}
\label{subsec:transfer}

Our dataset includes 1,000 carefully curated motion-transferred sequences, each derived from the source videos through various transformation scenarios. These sequences are specifically designed to evaluate different aspects of motion transfer capabilities, ranging from object transformations to comprehensive scene alterations. The transfers are organized into two primary categories, each addressing distinct challenges in motion transfer tasks:
\noindent\textbf{1. Cross-Category Transfers}

This category evaluates a method's capability to transfer motion patterns across different object categories while maintaining motion fidelity. The sequences are divided into three distinct transfer types:
\begin{figure}[t]
    \centering
    \begin{tabular}{c}
        \includegraphics[width=\linewidth]{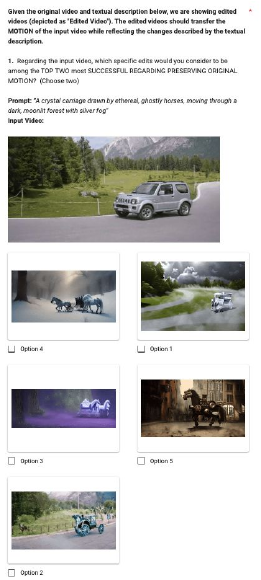} \\
    \end{tabular}
    \caption{\textbf{User Study Interface.} Given a reference video we ask for 3 different type of questions with 5 different options including DMT, MI, MD, VMC and MotionShop results.}
    \label{fig:user_study_interface}
\end{figure}

\noindent\textbf{Object-to-Object:} These transfers focus on motion preservation across different object categories while handling significant shape and appearance variations. Examples include:
\begin{itemize}
    \item Vehicle-to-creature transformations (e.g., car motion applied to a mechanical horse)
    \item Mechanical-to-organic conversions (e.g., robot movements mapped to flowing water)
    \item Rigid-to-deformable translations (e.g., toy motion adapted to cloth-like objects)
\end{itemize}
These sequences test the ability to maintain motion characteristics despite fundamental changes in object properties and physical constraints.

\noindent\textbf{Human-to-Character:} This subset specifically addresses the challenging task of transferring human motion to non-human characters while preserving natural movement patterns. Examples include:
\begin{itemize}
    \item Human dance movements applied to animated characters
    \item Sports motions transferred to fantasy creatures
    \item Gesture sequences mapped to mechanical entities
\end{itemize}
These transfers test the preservation of complex articulated motion while adapting to different skeletal structures and movement constraints.

\noindent\textbf{Animal-to-Object:} These sequences evaluate the transfer of organic motion patterns to inorganic objects, presenting unique challenges in motion adaptation. Examples include:
\begin{itemize}
    \item Bird flight patterns applied to flying vehicles
    \item Quadruped locomotion mapped to mechanical assemblies
\end{itemize}

\noindent\textbf{2. Scene Transformation Transfers}

This category focuses on evaluating motion preservation within dramatically altered environmental contexts, addressing two key aspects:

\noindent\textbf{Environment Changes:} These transfers test the ability to maintain motion fidelity while completely transforming the surrounding environment. The sequences include:
\begin{itemize}
    \item Context shifts (e.g., street scene to underwater environment)
    \item Scale transformations (e.g., human-scale to miniature worlds)
    \item Physical domain changes (e.g., terrestrial to aerial scenarios)
\end{itemize}
These sequences evaluate how well methods handle motion transfer when environmental physics and constraints change significantly.

\noindent\textbf{Style Transfers:} This subset focuses on artistic and stylistic transformations while maintaining motion integrity. Examples include:
\begin{itemize}
    \item Realistic to animated style conversions
    \item Contemporary to historical aesthetic adaptations
    \item Natural to fantastical scene transformations
\end{itemize}

Each transfer category is carefully designed to test specific aspects of motion transfer capabilities, from basic motion preservation to complex scene-level transformations. The sequences vary in complexity, duration, and transformation extent, providing a comprehensive evaluation framework for assessing motion transfer methods across different scenarios and applications. This structured approach enables systematic analysis of a method's strengths and limitations in handling various types of motion transfer challenges.

\subsection{Dataset Statistics}
\label{subsec:statistics}

Key characteristics of the dataset:
\begin{itemize}
    \item Resolution: 720×480 pixels
    \item Frame Rate: 15 FPS
    \item Duration: 1-7 seconds per video (Due to the frame processing limitation of CogVideoX)
    \item Total Frames: $\sim$45,000
    \item Format: MP4 (H.264 codec)
\end{itemize}

Here we note that the duration limit stems from the CogVideoX \cite{yang2024cogvideox} backbone. It can only process 49 frames at most. 

\subsection{Discussion}
\label{subsec:discussion}

MotionBench addresses several critical requirements essential for comprehensive motion transfer evaluation. First, it achieves comprehensiveness by encompassing a diverse range of motion types and scene compositions, ensuring broad coverage of real-world scenarios. Its scalable design provides sufficient data for meaningful model training and evaluation, while maintaining standardized evaluation protocols and metrics that enable consistent comparisons across different approaches. Furthermore, the dataset's inclusion of both synthetic and real-world scenarios ensures practical applicability across various use cases. Through these carefully considered design choices, MotionBench enables researchers to systematically compare motion transfer methods, analyze motion preservation capabilities, evaluate scene composition handling, and assess temporal consistency.

\end{document}